\documentclass[letterpaper, 10 pt, conference]{ieeeconf}  

\IEEEoverridecommandlockouts                              

\usepackage{times}
\usepackage{epsfig}
\usepackage{graphicx}
\usepackage{amsmath}
\usepackage{amssymb}
\usepackage{booktabs}
\usepackage[dvipsnames]{xcolor}
\usepackage{float}

\newcommand{\algoname}{\textit{EWareNet}}
\newcommand{\sota}{state-of-the-art}

\title{\LARGE \textbf{\algoname}:  Emotion-Aware Pedestrian Intent Prediction and Adaptive Spatial Profile Fusion for Social Robot Navigation }
\author{Venkatraman Narayanan, Bala Murali Manoghar, Rama Prashanth RV and Aniket Bera \\
{Department of Computer Science, Purdue University, USA}\\
}
\begin{document}
\maketitle
\thispagestyle{empty}
\pagestyle{empty}

\begin{abstract}

We present \textit{\algoname}, a novel intent and affect-aware social robot navigation algorithm among pedestrians. Our approach predicts the trajectory-based pedestrian intent from gait sequence, which is then used for intent-guided navigation taking into account social and proxemic constraints. We propose a transformer-based model that works on commodity RGB-D cameras mounted onto a moving robot. Our intent prediction routine is integrated into a mapless navigation scheme and makes no assumptions about the environment of pedestrian motion. Our navigation scheme consists of a novel obstacle profile representation methodology that is dynamically adjusted based on the pedestrian pose, intent, and affect. The navigation scheme is based on a reinforcement learning algorithm that takes pedestrian intent and robot's impact on pedestrian intent into consideration, in addition to the environmental configuration. We outperform current state-of-art algorithms for intent prediction from 3D gaits. 

\end{abstract}

\section{Introduction}
Recent technological advancements are making human-robot collaborations increasingly important. This can have many applications, including autonomous driving, social robotics, and surveillance systems. As humans and robots co-inhabit space, designing robots that follow collision-free paths and are socially acceptable to humans is becoming increasingly important, creating several challenges. For example, in the case of a densely crowded street, the robot needs to foresee the movements of an oblivious walker who is unaware they are in its path for friendlier navigation. Understanding the emotional perceptions in such scenarios enables the robot to make more informed decisions and navigate in a socially-conscious manner.

The study of human emotions has been a much-researched subject in areas like psychology, human-robot collaboration, interaction, etc. Certain studies have sought to identify the emotion in people based on cues such as body movement (walking style, etc.)~\cite{pose2emo,narayanan2020proxemo}, and verbal cues (speech tonalities and patterns)~\cite{speechemo,textemo}. There are also multimodal approaches that utilize a combination of these cues to recognize the person's emotion \cite{multiemo2,multiemo1, multiemo3}. Combining the information from the robot sensors and accounting for the uncertainty of the movements of oblivious walkers remains a significant challenge.

Several approaches have addressed a socially-acceptable robot navigation problem. Many recent works like ~\cite{ narayanan2020proxemo}, \cite{VEGA201972} focus on social robot navigation in crowded scenarios. However, these algorithms suffer from the following drawbacks:
\begin{enumerate}
    \item They rely on emotionally reactive navigation planning, which means that the perceived emotional/affective states are based entirely on historical gait sequences and do not consider gait predictions of the future.
    \item The influence of the robot's behavior on the emotional response of pedestrians is not taken into account.
    \item The proxemic constraints for pedestrian motion are heuristically computed from prior psychology studies. While this works in most cases, it does not efficiently capture each individual's uniqueness in comfort space, personal space, culture, etc.
\end{enumerate}

\begin{figure}[t!]
    \centering
    \includegraphics[width=\linewidth]{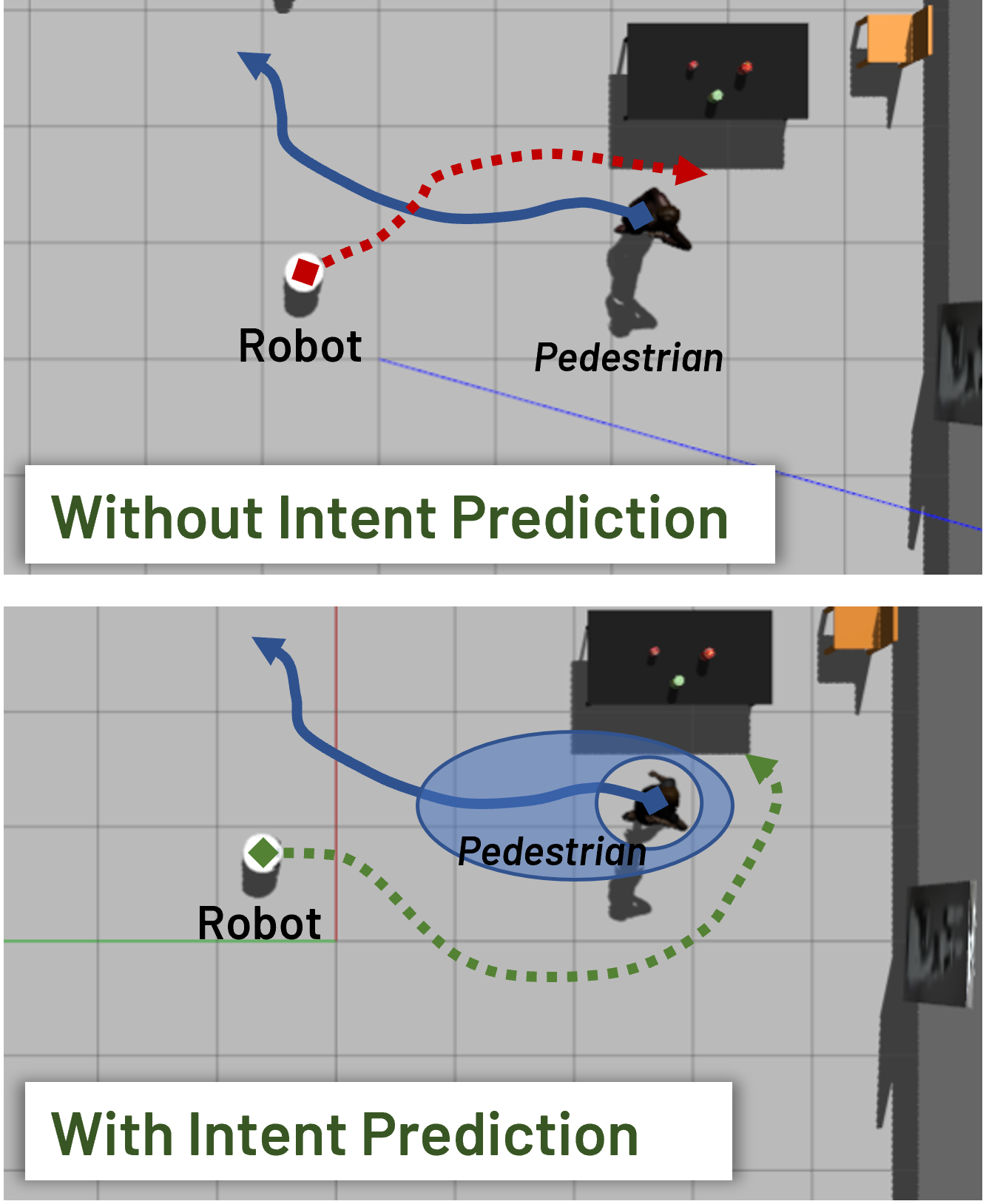}
        \caption{\textit{\textbf{EwareNet}:
        The top figure represents the path taken by the robot without intent prediction, where the trajectory intercepts the path to be taken by the pedestrian. The bottom figure shows the path taken by the robot with the intent prediction pipeline where even though there is enough space and a shorter route is available, the robot takes a longer path giving enough personal space.}
        \vspace{-0.5cm}}
    \label{fig:Emotion}
    \vspace{-0.1cm}
\end{figure}

To overcome these challenges,  we propose \algoname, a novel algorithm for a real-time, gait-based, intent-aware robot navigation algorithm in social scenes that takes into account the pedestrian's affect state, as depicted in figure \ref{fig:Emotion}. \algoname~is designed to work with commercial off-the-shelf RGB-D cameras and depth sensors that can be retrofitted to moving platforms or robots for navigation. The \textbf{major contributions} of our work can be summarized as follows:
\begin{itemize}
    \item We introduce a novel approach using \textit{transformers} to predict the full-body pedestrian intent or behavior in the form of trajectory/gaits based on historical gait sequences.
    \item We also present a novel navigation planning algorithm based on deep reinforcement learning that takes into consideration the environment, pedestrian intent, and the robot's reactionary impact on pedestrian behavior.
    \item Our method explicitly considers pedestrian behavior in crowds and the robot's impact on the pedestrians and environment. 
    \item Finally, we introduce an adaptive spatial density function to represent the proximal constraints for pedestrians
that captures pedestrians' unique personal comfort space in terms of pose, intent, and emotion.  
\end{itemize}

Pedestrian intent and affect (the scientific term for emotions) are very subjective and greatly influenced by many environmental and psychological factors. Therefore in this work, we focus only on the \emph{perceived} emotions from the point of an external observer as opposed to \emph{true} internal emotion. We also predict pedestrian intent as a potential full-body future trajectory based upon historical trajectories/gait patterns for a more efficient navigation strategy and may not represent actual/hidden intent. Additionally, we are only computing ``walking emotion'' as opposed to other forms of emotion \cite{montepare1987identification}.

The paper is organized as follows: Section \ref{Sec:Related_Work} presents related work, section \ref{sec:overview} gives an overview of our pipeline and describes each stage of our pipeline in detail, and finally, in \ref{sec:exp} we evaluate the results of our work.

\section{Related Work}
\label{Sec:Related_Work}
In this section, we present a brief overview of social-robot navigation algorithms. We also review related work on emotion modeling and classification from visual cues.

\subsection{Social Robotics and Emotionally-Guided Navigation}

A substantial amount of research focuses on identifying pedestrians' emotions based on body posture, movement, and other non-verbal cues. Ruiz-Garcia et al.~\cite{face2emotion1} and Tarnowski et al.~\cite{face2emotion2} use deep learning to classify different emotion categories from facial expressions.  \cite{multiemo3} use multiple modalities such as facial cues, pedestrian pose, and scene understanding. Randhavane et al.~\cite{randhavane2019identifying, randhavane2019liar} classify emotions into four classes based on affective features obtained from 3D skeletal poses extracted from pedestrian gait cycles. Their algorithm, however, requires a large number of 3D skeletal key points to detect emotions and is limited to single individual cases. Bera et al. \cite{bera2019emotionally,bera2020fg, bera2016glmp, bera2020you} classify emotions based on facial and body expressions along with a pedestrian trajectory obtained from overhead cameras. Although this technique accurately predicts emotions from trajectories and facial expressions, it explicitly requires overhead cameras in its pipeline. \cite{narayanan2020proxemo} provides an end-to-end deep learning-based emotion classification approach that takes in skeletal gaits from an arbitrary view.  \cite{dorbala2020can} modeled the navigation of a social robot based on human-robot trust interactions.

\subsection{Intention Prediction}
In social robotics, an accurate prediction of pedestrian trajectories plays a significant role in robot decisions in terms of reactive response, navigation planning, etc. For simplicity and ease of computation for navigational robots, the problem of pedestrian intention prediction is largely modeled as a trajectory prediction problem. Recurrent Neural Network based architectures are widely used for predicting pedestrian trajectories~\cite{rasouli2019pie, pavllo2018quaternet, abuduweili2019adaptable}.~\cite{mohamed2020social} use a Graph Convolution method to predict the trajectories and showed that Temporal Graph Convolutions are much better at predicting pedestrian trajectories compared to recurrent networks.


Most recurrent networks for trajectory prediction leverages some form of attention mechanism to improve the trajectories. \cite{abuduweili2019adaptable, rasouli2019pie} uses a history of observed trajectories with either predicted future trajectory or location around pedestrians to predict the intent. \cite{pavllo2018quaternet} provide skeletal joint kinematics as attention for trajectory prediction. These models, one way or the other, solely depend on the trajectory and kinematics of joints for predicting the intent, and there is no attention given to the emotional state of the pedestrians.

\subsection{Proxemic Constraints Modeling}

Usually, robots working in pedestrian environments have used navigation algorithms where all obstacles are considered of similar relevance, including people. \cite{bera2019emotionally} a have studied the effects of comfort space of pedestrians from a psychological perspective and showed that comfort space varies based on the emotion of pedestrians. To avoid discomforting pedestrians, social robots must consider unique entities, evaluating the people’s level of comfort with respect to the route of the robot.
The navigation model of \cite{bera2019emotionally, narayanan2020proxemo} uses the predicted emotions and uses a constant multiplier to maintain proper comfort space with the pedestrians while planning the robot path.

Vega et al. \cite{VEGA201972} proposed a pedestrian-aware navigation strategy based on space affordances. The method is built upon using an adaptive spatial density function that efficiently clusters groups of people according to their spatial arrangement. The paper \cite{li2020facial} discusses how an agent should learn the behavior from a reward provided by a live human trainer rather than the usual pre-coded reward function in a reinforcement learning framework. With the designed CNN-RNN model, our analysis shows that telling trainers to use facial expressions and competition can improve accuracy for estimating positive and negative feedback using facial expressions.

All these methods consider emotion as a constant metric for formulating proxemic constraints and do not adapt the navigation strategy to the change in the emotion of pedestrians.

\section{Overview and Methodology}
\label{sec:overview}
{\algoname} is a novel algorithm for intent-aware and socially-acceptable navigation through crowded scenarios. The pedestrian intentions are predicted based on 3D skeletal trajectories from an onboard RGB-D camera. Our navigation algorithm then uses these trajectories for socially-acceptable navigation. Our navigation planning is adaptive to the individual comfort space constraints of pedestrians and to the uncertainty in the sensor suite of the robot. Based on a Gaussian distribution, we use dynamic obstacle representation for pedestrians to capture individuality in comfort space constraints.

The following subsections will describe our approach in detail. We discuss the details of the datasets used for training our algorithm, along with the pre-processing techniques employed. Following this, we focus on our intent prediction routine, where we also discuss our pedestrian pose extraction from an RGB-D camera briefly. Finally, we discuss our navigation algorithm.

\begin{figure}[h]
\centering
\includegraphics[width=0.5\textwidth]{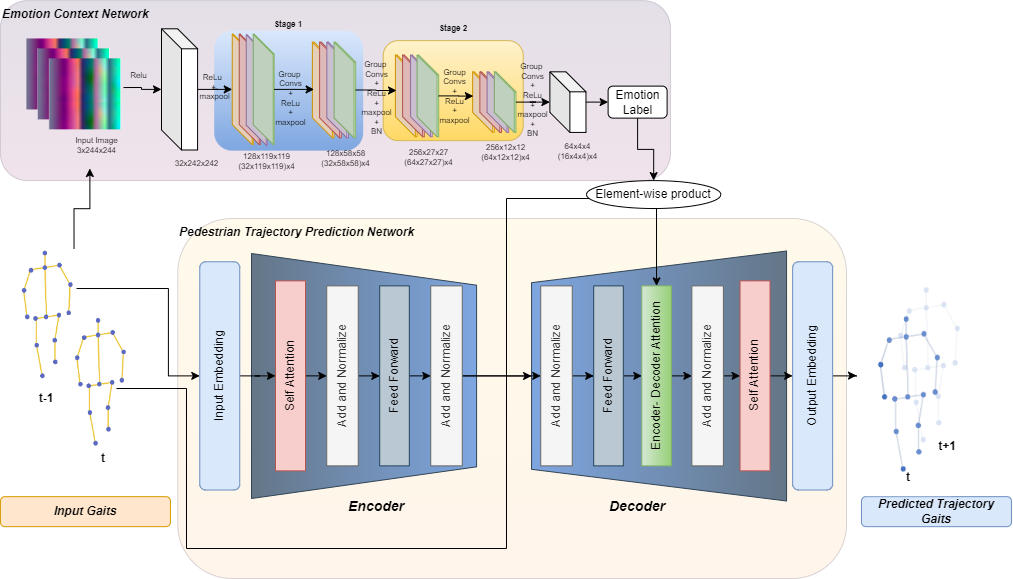}
\textit{ \caption{\textbf{Pedestrian Intent Prediction}: Our proposed method for predicting pedestrian intentions as full-body skeletal trajectory paths. It consists of a two-part network - \textcolor{DarkOrchid}{\textbf{Emotion Context Network}} and \textcolor{Dandelion}{\textbf{Pedestrian Trajectory Prediction Network}}. The \textit{Emotion Context Network} provides an additional \textbf{attention mechanism} on top of the self-attention mechanism in transformer networks.}}
\label{fig:ewarenet}
\end{figure}

 \subsection{Pedestrian Pose Extraction}
 
The primary objective of our work involves the effective use of pedestrian temporal skeletal poses for more harmonious navigation of a robot in crowded situations. We rely on the system, described in \cite{dabral2018learning}, to extract the poses from pedestrians walking in a crowded real-world scenario from an RGB-D camera. The system consists of a two-step network trained in a weakly supervised fashion. A \textit{Structure-Aware PoseNet (SAP-Net)} provides an initial pose estimate based on spatial information of joint locations of people in the video. Later, a time-series \textit{Temporal PoseNet (TP-Net)} corrects the initial estimate by adjusting impermissible joint angles and joint distances (based on human physiology and geometry restrictions). The temporal network also helps in tracking the pedestrian individual across the video frames. Hence, we have a set of temporally correct, \textit{Pedestrian Joint Pose Sequence} (henceforth referred to as \textit{gaits}) for every pedestrian in the video as an output from the system.

\begin{figure}[h]
\centering
\includegraphics[width=0.485\textwidth]{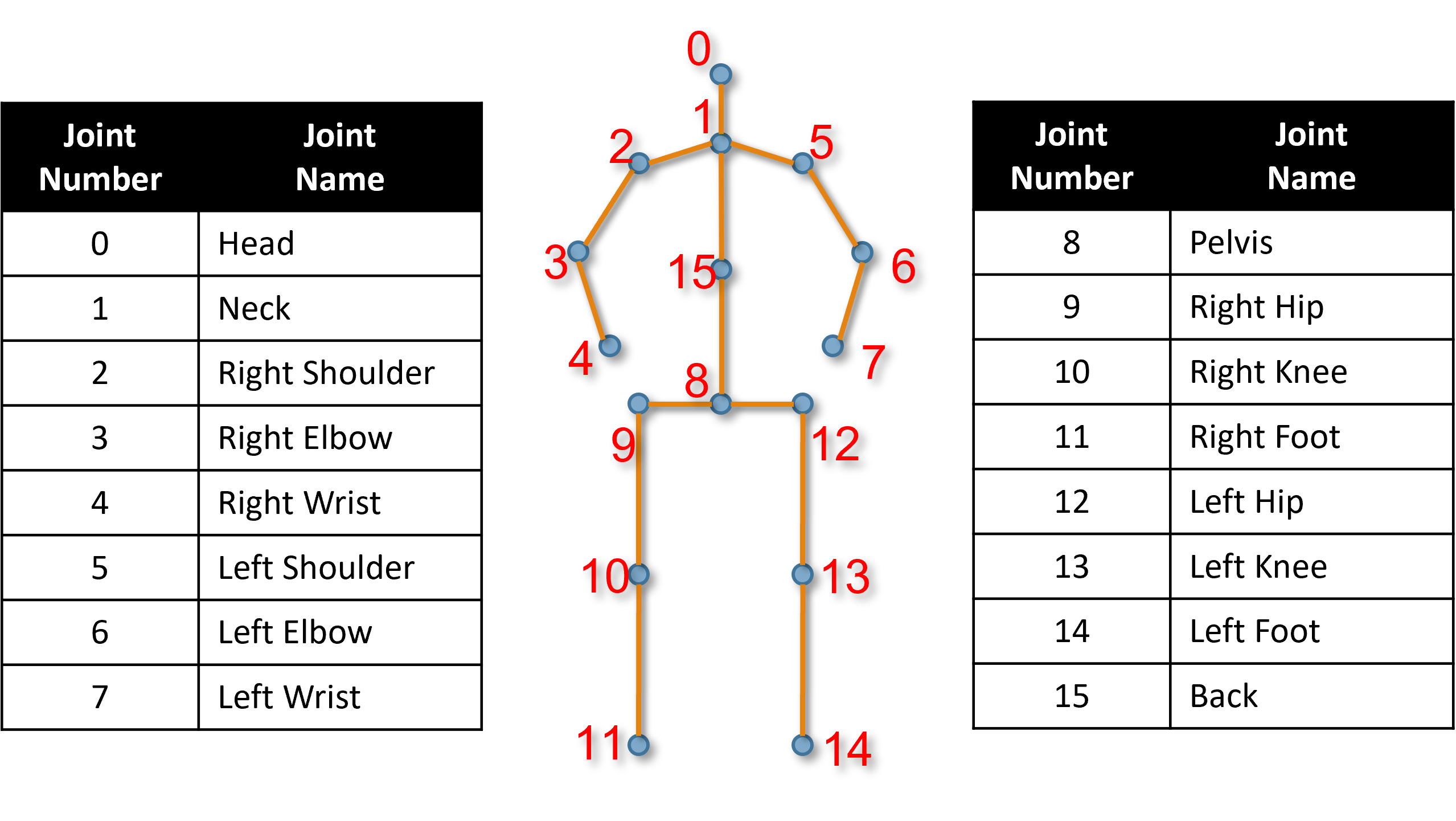}
\caption{\textit{\textbf{Skeleton Representation}: We represent a pedestrian by $16$ joints ($j_i$). The overall pose of the pedestrian is defined using these joint positions.}
}
\label{fig:skeleton_temporal}
\vspace{-10pt}
\end{figure}

Similar to \cite{narayanan2020proxemo}, our \textit{Pedestrian Pose Extraction} network extracts a representation for the pedestrians as $16$ skeletal joints as shown in figure \ref{fig:skeleton_temporal}. Every pose, $P \in \mathbb{R}^{16\times3}$ of a pedestrian consists a set of 3D positions of each joint $j_i$, where $i \in \{0,1, ..., 15\}$. For any RGB-D video $V$, we represent the gait extracted using 3D pose estimation as $G$. The gait $G_i,t$ is a set of 3D poses for pedestrian $i$ over ${P_1, P_2,..., P_t}$ where $t$ is the frame number of the input video $V$.
 
 \subsection{Pedestrian Intent Prediction} \label{emotionmech}
Our Pedestrian Intent Prediction module is designed as a trajectory prediction system based on a set of skeletal gait sequences. The intent prediction system consists of three parts, \textit{(i) an Emotion Context Network (ECN), (ii) Pedestrian Trajectory Forecasting Network}. Figure \ref{fig:ewarenet} provides a schematic representation of our Pedestrian Intent Prediction algorithm.

\subsubsection{Emotion Context Network}\label{emotionRec}
Our emotion context network extends on~\cite{narayanan2020proxemo}. It is a multi-view emotion classification network. The architecture takes in the gait sequences consisting of $16$ joints (depicted in figure \ref{fig:ewarenet}) for $75$ time steps to classify them into one of $4$ emotions \textit{(Happy, Sad, Angry, Neutral)}. The gait sequences are processed into image embeddings to leverage faster processing abilities of \textit{Convolution Neural Networks}~\cite{narayanan2020proxemo}. The embedded images are fed to the classification network. The classification network \textit{(ECN)} consists of multiple layers of grouped convolutions—the grouped convolutions aid in classifying the emotions for different camera angle view groups. The extracted emotion is fed as an attention mechanism to our \textit{Transformer Network} to obtain intent-aware trajectory predictions.
 \begin{equation}
 \label{eq1}
     e_i = ECN(I_i)
 \end{equation}
Each gait sequences \textit{(i)} is assigned an emotion label, $(e_i)$, based on equation \ref{eq1}, where $ECN$ represents the emotion classification network, $I_i$, represents the image embedding of the gait sequences $i$
 \subsubsection{Pedestrian Trajectory Forecasting Network}
 The Pedestrian Trajectory Forecasting network predicts the intent-aware future full-body gait trajectories for every pedestrian based on past gaits. Our Pedestrian Trajectory Forecasting network is inspired by transformer networks depicted in \cite{giuliari2020transformer}. Transformer networks have successfully produced \sota \ results involving time-series sequences. Wen et al. \cite{Wen2022TransformersIT} conducted extensive analysis, which concluded that Transformers are superior in modeling long-range dependencies, training can be parallelized, and does not suffer from vanishing gradients when compared to the recurrent counterpart approaches. They have been extensively used in NLP tasks~\cite{devlin2018bert,brown2005language} and time-series forecasting~\cite{lim2019temporal}. Guiliari et al. ~\cite{giuliari2020transformer} uses transformer networks to generate \sota \ results on pedestrian trajectory prediction. 
 \vspace{-10pt}
\begin{equation}
    \label{eqn2}
    e_{i, t} = {G_{i, t}}^T . W_g
\end{equation}
 
The input gait sequences for each pedestrian, $G_{i, t}$,  are fed to a linear encoding layer given by equation \ref{eqn2}. The embedded gaits inputs are provided to the transformer network. Transformer (TF) network is a modular architecture (figure \ref{fig:ewarenet}). The transformer network follows an encoder-decoder style network, each composed of $6$ layers, with three building blocks each: (i) a self-attention module, (ii) a feed-forward fully-connected module, and (iii) two residual connections after each of the previous blocks. The self-attention modules within the TF net provide the capability to capture non-linearities in the trajectory.
 
The encoder model creates a representation, ${E_s}^{(i, t)} $, of the input gait sequences, forming a memory module. The encoder representation is utilized by the decoder model to auto-regressively predicts future trajectories. 
 
 The \textit{encoder-decoder attention} layer of the decoder network is fed with encoder representation and emotion context from emotion context network in \ref{emotionmech}. The emotion probabilities, $e_i$, from the \textit{ECN} are multiplied element-wise with the encoder outputs to be fed as attention to the decoder. For simplicity, we assume that the emotion context remains constant for all the timesteps predicted by the decoder. The intent-aware trajectory predictions are used in our navigation algorithm.
 
 \subsection{Proxemic Contraints Modeling}
 The personal space constraints are derived from prior works in psychology~\cite{ruggiero2017effect}. Personal space constraints are essential in social robotics and emotion-guided navigation ~\cite{kirby2010social}. The comfort space constraints defined in~\cite{social5}, \cite{ruggiero2017effect}~describe the social norms that the robot must consider not to cause discomfort to people around it. The comfort space constraints related in \cite{ruggiero2017effect}, ~\cite{social5}, \cite{narayanan2020proxemo} do not consider the individuality of pedestrians. On the other hand, our approach is able to capture the individuality of the comfort space for pedestrians. 
 
 Our pedestrian personal space modeling is an adaptive spatial density function, similar to~\cite{VEGA201972}. In an arbitrary global map in space, a pedestrian, \textit{i},  is represented by position, orientation, and emotion (from \ref{emotionRec}), $h_i = {(x_i, y_i, \theta_i, C_i)}^T$. Here $(x_i, y_i)$ represents the position of the pedestrian, $\theta_i$ represents the orientation, and $C_i$ represents the comfort space derived from emotion. The position coordinates from each pedestrian will be derived from the LiDAR coordinates after mapping detected pedestrians from the camera frame to the LiDAR frame using projective geometry. The orientation and emotion are derived from section \ref{emotionRec}. The view-group angle predicted by the model is used as the approximated orientation. The gaussian expression for personal space is defined by equation \ref{eqn:comfortgauss}.
 \begin{equation}
    \label{eqn:comfortgauss}
    \begin{split}
g_{h_i} &= C_i*\exp(-k_1{(x-x_i)}^2 \\
&+k_2(x-x_i)(y-y_i) \\
&+k_3(y-y_i)^2)
    \end{split}
\end{equation}
Here, $C_i$ is the comfort space multiplier given by,
  \vspace{-7pt}
\begin{align}\label{eqn:socdist}
C_i = \frac{\sum_{j=1}^{4} c_j \cdot \max{(e_j)} }{\sum_{j=1}^{4}{e_j}} \cdot v_g
\end{align}
where $e_{j}$ represents a column vector of the emotion context output from section \ref{emotionRec}, which corresponds to the group outcomes for each individual emotion. $c_j$ is an individual's comfort space constant derived from psychological experiments described in~\cite{ruggiero2017effect}, chosen from a set $\{90.04, 112.71, 99.75, 92.03\}$ corresponding to the comfort spaces (radius in cm) for \{\textit{happy, sad, angry, neutral}\} respectively. These distances are based on how comfortable pedestrians are while interacting with others. $v_g$ is a view-group constant chosen from a set of $\{1, 0.5, 0, 0.5\}$ based on the view group $g$, defined in \cite{narayanan2020proxemo}. 

Also, $k_1, k_2, k_3$ in equation \ref{eqn:comfortgauss} are the coefficients taking into account the orientation $\theta_i \in [0, 2\pi)$, defined in equations \ref{eqn:k1}, \ref{eqn:k2}, \ref{eqn:k3}.
 \vspace{-10pt}
\begin{align}
    \label{eqn:k1}
    k_1(\theta_i) = \frac{cos({\theta_i}^2)}{2\sigma^2} + \frac{sin({\theta_i}^2)}{2\sigma_s^2}
\end{align}
 \vspace{-10pt}
\begin{align}
    \label{eqn:k2}
    k_2(\theta_i) = \frac{sin(2\theta_i)}{4\sigma^2}  \frac{sin(2\theta_i)}{4\sigma_s^2}
\end{align}
 \vspace{-10pt}
\begin{align}
    \label{eqn:k3}
    k_1(\theta_i) = \frac{sin({\theta_i}^2)}{2\sigma^2} + \frac{cos({\theta_i}^2)}{2\sigma_s^2}
\end{align}
where $\sigma_s$ is the variance to the sides of the pedestrian ($\theta_i \pm \frac{\pi}{2}$) and $\sigma$ is the variance in the direction of the pedestrian orientation.

We understand that personal space distances depend on many factors, including cultural differences, the environment, or a pedestrian's personality. Hence we have modeled the personal space distances as an adaptive Gaussian function that is dynamically updated by our navigation algorithm (detailed in section \ref{sec:navigation}).
 
\subsection{Intent-Aware Navigation}
\label{sec:navigation}

Our navigation framework is defined as a policy network trained to reach the goal while adjusting to the obstacles and personal space constraints of the pedestrians. Three consecutive LiDAR scans and tracked human poses are processed to accommodate our adaptive proxemic constraints for each detected pedestrian. Our policy network is inspired by ~\cite{yang2019embodied,fan2018crowdmove}. Our policy network receives the processed LIDAR scans and the intent-aware trajectory predictions for each pedestrian and outputs probabilities over the action space considered for the navigation task. 

\textbf{State Space:} As discussed earlier, we consider three consecutive liar frames and tracked human poses. Along with these two components, the relative goal position and the robot's current velocity components constitute the state space. Each LIDAR scan is represented as $l_i \in \mathbb{R}^{512}$ and each pose is represented as $h_j \in \mathbb{R}^{16}$. Thus the state space is given by equation \ref{eq:state}
 \vspace{-7pt}
\begin{align}
\label{eq:state}
s_t = \{l_0, l_1, l_2, h_0, h_1, ..., h_i\}
\end{align}

Since we consider three consecutive LIDAR scans and 18 consecutive traced skeleton poses, the state space is represented as $s_t \in \mathbb{R}^{(3\times512)\times(18\times16)}$

\textbf{Action Space:} The action space is a continuous set of permissible
velocities. The action of the robot includes translational and rotational velocity. To accommodate the robot kinematics, we set the bounds for translational velocity, $v \in [0.0, 1.0]$, and rotational velocity, $w \in [-1.0, 1.0]$. We sample actions at step $t$ using equation \ref{eq:act}.

\begin{figure}[h]
\centering
\includegraphics[width=0.5\textwidth]{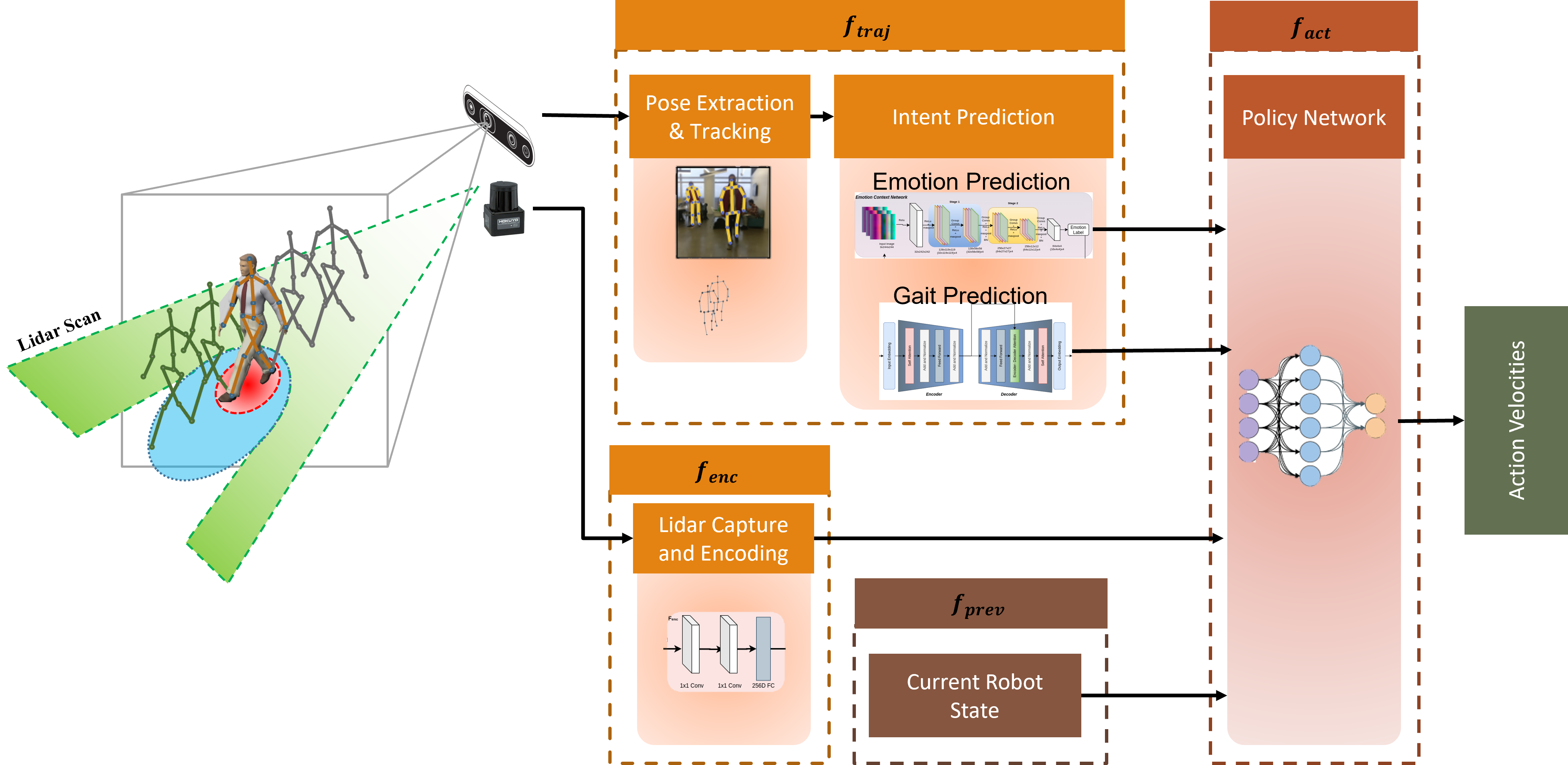}
\caption{\textit{Our Navigation algorithm takes in the intent-aware pedestrian trajectories, LiDAR scans, and robot's positional information as input to a policy network. The policy network outputs the \textit{robot velocity vectors} as action space.}}
\label{fig:rlnet}
 \vspace{-5pt}
\end{figure}
 \vspace{-15pt}
\begin{align}
\label{eq:act}
a_t = \pi(l_0, l_1, l_2, h_0, h_1, ..., h_i)
\end{align}
where $l_0, l_1, l_2$ represent the three consecutive processed LiDAR scan frames and $h_0, h_1, ..h_t$ represent the historical and predicted pedestrian trajectories concatenated together.

\textbf{Policy Network}: The policy network has four sub-component networks $\{f_{traj}, f_{enc}, f_{prev}, f_{act}\}$ (depicted in figure \ref{fig:rlnet}), where $f_{traj}$ depicts our Pedestrian Intent Prediction network to encode the pedestrian trajectories, $f_{enc}$ depicts our network to encode LiDAR frames from the robot, and finally $f_{act}$ represents the network that takes the above-mentioned encodings along with robot's previous state and velocity to predict the action velocities to the robot. We embed the concatenated (historical \& predicted) trajectories to skeletal gait to image embedding in~\cite{narayanan2020proxemo}, resize them to $244 \times 244$, and pass them to $f_{traj}$, which consists of four $5\times5$ Conv, BatchNorm, ReLU. Each Conv block is followed by a $2\times2$ MaxPool blocks, producing an encoded image of pedestrian trajectories $ z^{img}_{t} = f_{traj}([h_0, h_1, h_i])$

The three consecutive lidar frames are passed through two $1\times1$ Conv, followed by a $256D$ fully-connected (FC) layer. The lidar frames are encoded as $z^{lidar}_{t} = f_{enc}([l_0, l_1, l_2])$. The $f_{act}$ is a multi-layer perceptron (MLP) network, with 1 $128D$ FC hidden layer, and finally produces the action velocities. $f_{act}$ takes in the encoded pedestrian trajectories, $z^{img}_{t}$, lidar encodings, $z^{lidar}_{t}$, previous velocity, $v_{t-1}$, goal position, $s_g$, and current robot position, $s_t$,  to predict the robot velocities at time $t$, given by equation \ref{eq:act1}.
 \vspace{-7pt}
\begin{align}[h]
\label{eq:act1}
v_t = f_{act}([z^{img}_{t}, z^{lidar}_{t}, v_{t-1}, s_g, s_t])
\end{align}

$v_t$ is then sent to a linear layer with softmax to derive the probability distribution over the action space, from which the action is sampled. We learn $\{f_{traj}, f_{enc}, f_{act}\}$ via reinforcement learning. 

\textbf{Rewards}: Our reward function for the policy network extends \cite{fan2018crowdmove}. Our goal is to find a good strategy to avoid collisions during navigation, minimize the arrival time and avoid any negative impact on the pedestrians while navigating, i.e., avoid pedestrian emotions progressing into ${Angry, Sad}$ while navigating (unless respective humans already display negative emotions. The reward function to achieve the mentioned goals is given in equation \ref{eqn:overallreward}.
 \vspace{-5pt}
\begin{equation}
    \label{eqn:overallreward}
    r^t = r_g^t + r_c^t + r_w^t + r_e^t
\end{equation}
The reward $r$ at time $t$ is a combination of reward for reaching goal, $r_g$, reward for avoiding collisions, $r_c$, reward for smooth movement, $r_w$ and reward for avoiding negative impact on pedestrians, $r_e$. The reward for reaching the intended goal/target is given by equation \ref{eqn:goalrew}.
 \vspace{-5pt}
\begin{equation}
    \label{eqn:goalrew}
    r_g^t = \begin{cases}
            r_{arrival}, & \text{if $||p^t - g|| < \xi$}.\\
            w_g(||p^{t-1} - g|| - ||p^t - g||), & \text{Otherwise}.
            \end{cases}
\end{equation}
The penalty for colliding with obstacles is given by equation \ref{eqn:penalrew}.

\begin{equation}
    \label{eqn:penalrew}
    r_c^t = \begin{cases}
            r_{collision}, & \text{if robot collides}.\\
            0, & \text{Otherwise}.
            \end{cases}
\end{equation}
For ensuring smooth navigation, the penalty for large rotational velocities is given by equation \ref{eqn:smoothrew}.

\begin{equation}
    \label{eqn:smoothrew}
    r_w^t = \begin{cases}
            w_w|w^t|, & \text{if $|w^t| > 0.7$}.\\
            0, & \text{Otherwise}.
            \end{cases}
\end{equation}
To ensure the robot does not alarm any pedestrian, we penalize the robot for any change in detected pedestrian emotion into negative emotions, ${Angry, Sad}$, with equation \ref{eqn:emorew}. The penalty is only applied when the pedestrian emotion, $e_i$, changes from non-negative emotion to negative emotion. In other words, we do not penalize the robot when pedestrian emotion remains unchanged.

\begin{equation}
    \label{eqn:emorew}
    r_e^t = \begin{cases}
            r_e, & \text{if $e_0, .., e_i \in \{Angry, Sad\}$}.\\
            0, & \text{Otherwise}.
            \end{cases}
\end{equation}

In our implementation, we use $r_{arrival} = 15, w_g = 2.5, r_{collision} = -15, w_w = -0.1, r_e = -2.5, \xi = 0.1$

\section{Experiments and Results}
\label{sec:exp}
\subsection{Pedestrian Intent Prediction}
\subsubsection{Metric}
\label{sec:intmetrics}
Since our pedestrian intent prediction is modeled as a trajectory prediction module, we evaluate the network with metrics commonly used for full-body trajectory/pose prediction networks. We use Mean Squared Error (MSE), which is the average $l_2$ distance between the ground-truth and the predicted poses at each time step for all pedestrians present in the frame.

\subsubsection{Datasets}\label{sec:intentdata}
 In this work, we leverage datasets designed for specific tasks of our approach.
 
 \textbf{Pedestrian Emotion Datasets}: Our intent prediction system is reinforced with emotion labels extracted from the historical gaits. For this purpose, we use two labeled datasets by Randhavane et al.~\cite{Ewalk} and Bhattacharya et al.~\cite{bhattacharya2020step}. It contains 3D skeletal joints of  $342$ and $1835$ pedestrian gait cycles each ($2177$ gait samples in total).
 
 \textbf{Pedestrian Intent Prediction Datasets}: Our intent prediction system is modeled as a trajectory prediction system based on past trajectory/gait sequences reinforced with pedestrian emotions for the same. We use two social datasets, created from NTU RGB+D 60 ~\cite{dataset:NTURGBD} and PoseTrack ~\cite{dataset:PoseTrack} for training and evaluating our and demonstrate its superior performance against several relevant baselines. The NTU dataset contains both single-person actions and mutual actions. The PoseTrack is a large-scale multi-person dataset based on the MPII Multi-Person benchmark. The
dataset covers a diverse variety of interactions, including
person-person and person-object, in crowded, dynamic scenarios. For our experiments, we consider only the actions involving pedestrians \textit{walking and/or running}. Hence we only select samples that fall under these categories.

\subsubsection{Implementation Details} 
For training, our dataset (\ref{sec:intentdata}) has a train-validation split of 90\%-10\%. 
We perform training using an ADAM \cite{kingma2014adam} optimizer, with decay parameters of ($\beta_{1} = 0.9 $ and $\beta_{2}= 0.999$). The experiments were run with a learning rate of $0.009$ and with $10\%$ decay every 250 epochs. The models were trained with MSE loss, \textbf{$\mathcal{L}$}.
The training was done on 2 Nvidia RTX 2080 Ti GPUs having 11GB of GPU memory each and 64 GB of RAM.

\subsection{Intent-Aware Navigation}
\subsubsection{Metric}
\label{Sec:metrics}
We evaluate our navigation algorithm using three metrics based on prior literature 
(i) Total distance traveled from source to goal, (ii) Time taken to reach the goal, (iii) Average personal space deviation $(\Delta_{ps}^{avg})$, i.e., the difference between expected personal space and actual personal space provided by the robot while passing the pedestrian averaged across all pedestrian encounters. 

\subsubsection{Implementation Details}
\label{sec:impdet}
\begin{figure}[ht]
\centering
\includegraphics[width=0.5\textwidth]{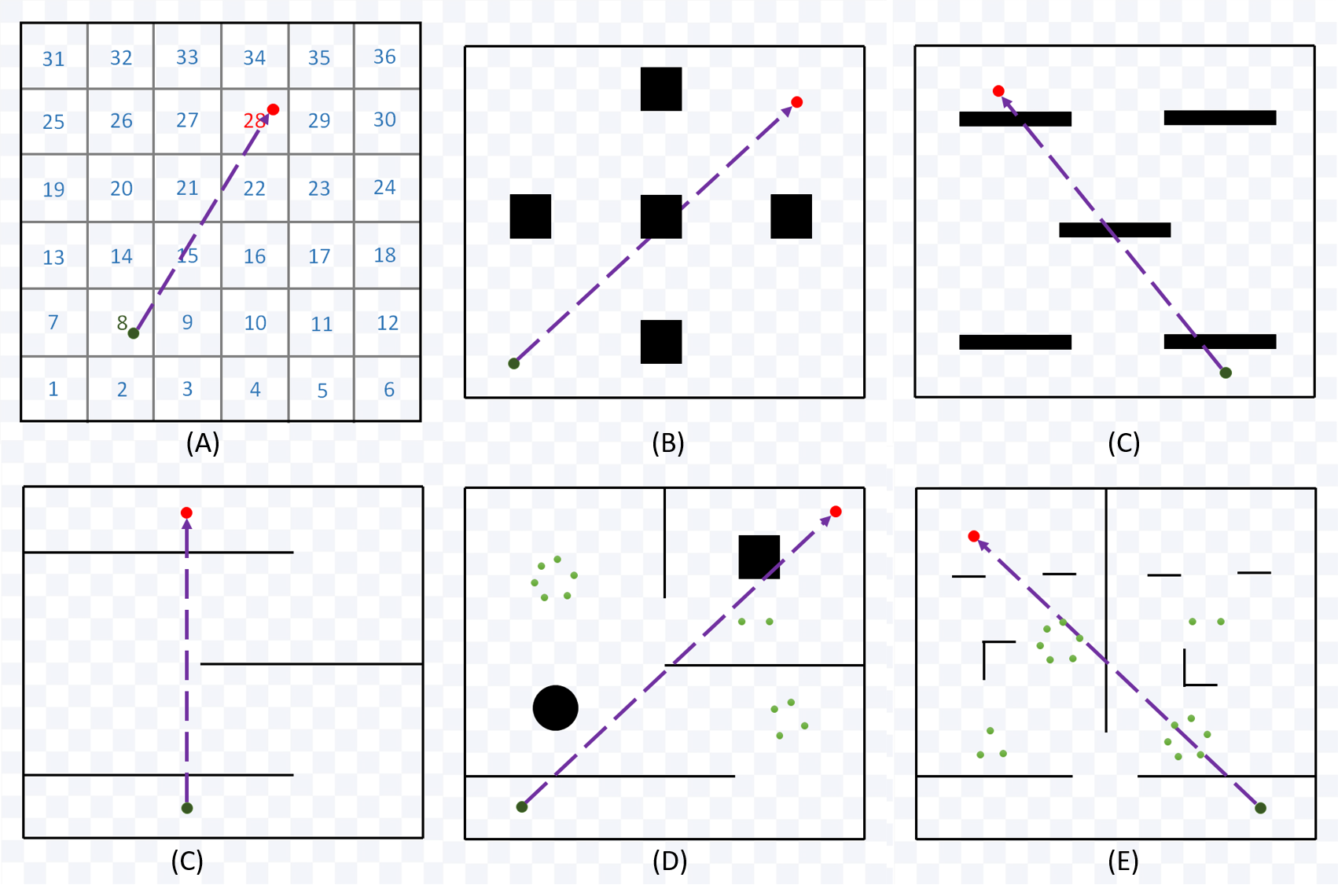}
\caption{\textit{The various scenarios used to train our navigation algorithm. As mentioned in section \ref{sec:impdet}, several configurations of parameters within a scenario are tweaked to generate multiple training/testing scenarios based on figures A-E.}}
\label{fig:tain}
\end{figure}

We rely on simulation environments to train our policy network. We use Stage Mobile Robot Simulator \cite{vaughan2008massively} to generate multiple scenarios (see in figure \ref{fig:tain}) with obstacles to train our policy network. We perform a multi-stage training of our policy network to aid policy convergence. We initially train the network on random scenarios from figure \ref{fig:tain} - (A), without any obstacles. The map is divided into a $6 \times 6$ grid. A random point from two different grids is chosen as the source and target location for training the random policy. Once the robot converges with a policy to reach a goal without obstacles, we further train on random samples with obstacles. Sample maps are shown in figure \ref{fig:tain} - (B to E). Furthermore, we place a random number of pedestrians, ranging anywhere from 5 to 50, with emotions and trajectories from the datasets by Randhavane et al.~\cite{Ewalk}, and Bhattacharya et al.~\cite{bhattacharya2020step} for training our policy network. Thus, we train the robot to navigate with pedestrian proxemic constraints using RMSProp \cite{hinton2012neural} with a learning rate of $0.00004$ and $\epsilon = 0.00005$.

\subsubsection{Analysis on Experiment Results}
\begin{figure}[h]
\centering
\includegraphics[width=0.5\textwidth]{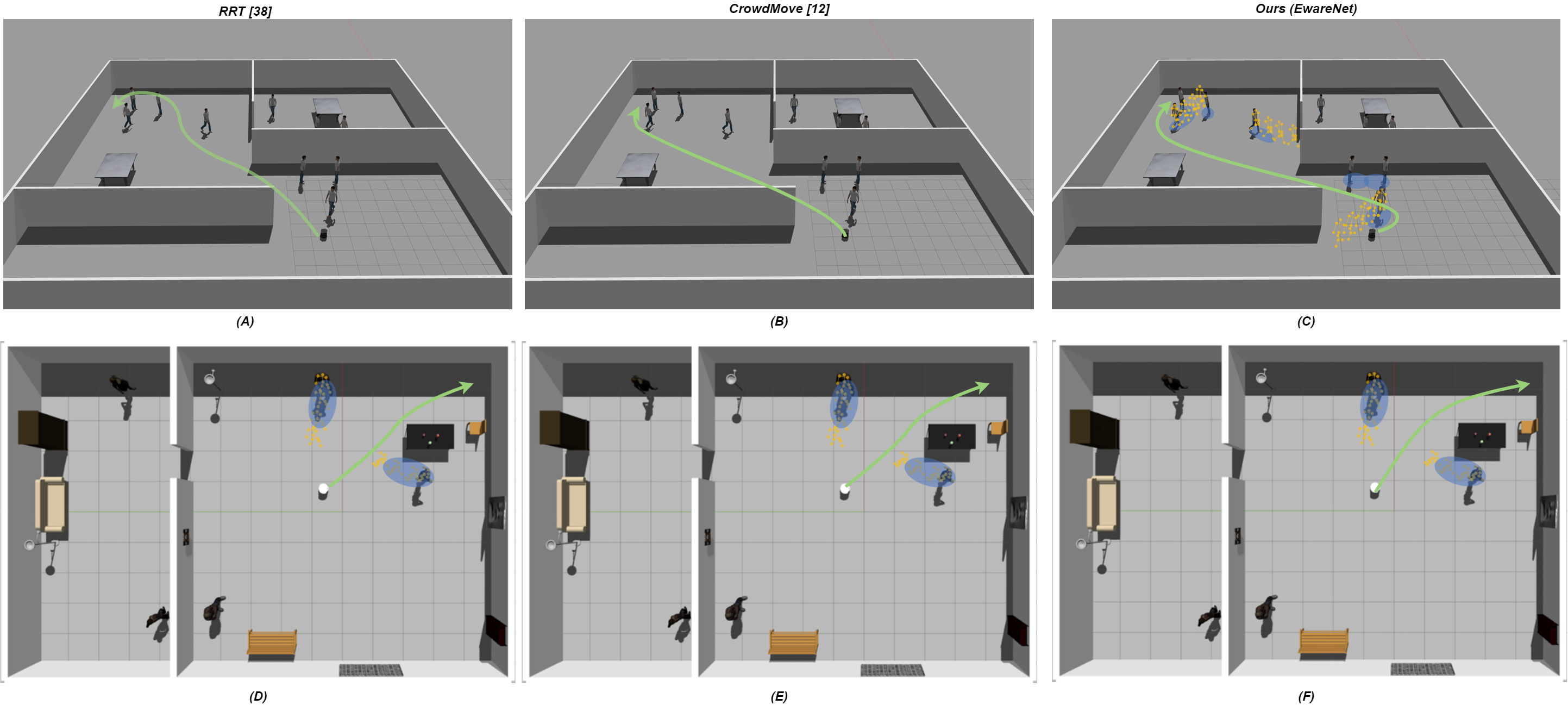}
\caption{\textit{\textbf{Experimental Demo}: We showcase the performance of our {\algoname} with other algorithms mentioned in Table \ref{tab:navres}. Here the trajectory taken by the robot is depicted in \textcolor{green}{Green}, along with intent-aware trajectories as gaits in \textcolor{yellow}{Yellow} and adaptive personal/comfort space area around a pedestrian in \textcolor{blue}{Blue}. Figures A \& D represent the experiment with RRT \cite{VEGA201972}, B \& E represent the experiment with CrowdMove \cite{fan2018crowdmove} and C \& F represents {\algoname}.}}
\label{fig:demo}
\vspace{-10pt}
\end{figure}

We tabulate the results from our experiment in Table \ref{tab:navres}. We compare our implementation with \cite{fan2018crowdmove} and \cite{VEGA201972}. We use scenarios $5$ and $6$ from figure \ref{fig:tain} to perform our comparison studies. We also perform $20$ iterations of the testing for each scenario under multiple environmental configurations (initial and goal positions, number and position of obstacles, pedestrian and emotion parameters). It can be seen that our implementation has the best Average Personal Space Deviation across all our experiments. Figure \ref{fig:demo} showcases the experimental results of our comparison studies.

\begin{table}[tb]
\centering
\vspace{0.25cm}
\resizebox{\columnwidth}{!}{
\begin{tabular}{@{}|c|c|c|c|@{}}
\toprule
\hline \hline
Method  & Total Distance (m) & Avg. Time (min) & $\Delta_{ps}^{avg}$ (m)\\ \hline
\multicolumn{4}{c}{Scenario 5} \\ \hline
Crowdmove \cite{fan2018crowdmove} & 152.5 & 2.6 & 25.3 \\ \hline
RRT \cite{VEGA201972} & 210.4 & 3.5 & 27.4 \\ \hline
PRM \cite{VEGA201972} & 217.6 & 3.6 & 26.8 \\ \hline
\textbf{Ours} & 172.3 & 2.9 & \textbf{10.2} \\ \hline
\multicolumn{4}{c}{Scenario 6} \\ \hline
Crowdmove \cite{fan2018crowdmove} & 224.5 & 7.2 & 57.8 \\ \hline
RRT \cite{VEGA201972} & 262.8 & 9.5 & 63.4 \\ \hline
PRM \cite{VEGA201972} & 245.6 & 8.7 & 61.8 \\ \hline
\textbf{Ours}  & 237.6 & 8.7 & \textbf{32.8} \\ \hline
\bottomrule
\end{tabular}
}
\caption{Comparison of {\algoname} with various baselines using scenarios $5$ and $6$ from figure \ref{fig:tain}. We report the metrics mentioned in section \ref{Sec:metrics} compared with \cite{fan2018crowdmove} and \cite{VEGA201972}. We can see that our {\algoname} has the \textbf{best \textit{Average personal space deviation}} ($\Delta_{ps}^{avg}$) across all our experiments.}
\label{tab:navres}
\vspace{-20pt}
\end{table}

\section{Conclusion and Future Work}
We introduce a novel approach using \textit{Transformer Attention Networks} to predict the full-body human intent or behavior in the form of trajectory/gaits based on historical gait sequences. We also present a novel navigation planning algorithm based on deep reinforcement learning that takes into consideration the environment, human intent, and the robot's reactionary impact on human behavior. Our method explicitly considers pedestrian behavior in crowds and the robot's impact on the environment and the uncertainty in the robot's sensor suite. Finally, our adaptive spatial density function to represent the proximal constraints for pedestrians captures their unique, personal comfort space in terms of pose, intent, and emotion. In the future, we plan to develop a platform to model pedestrian gaits that embody emotion to help benchmark emotion-guided social robot navigation algorithms. We also intend to extend our work to include other markers of emotion (such as facial expressions), specifically to address static pedestrians.

{\small
\bibliographystyle{IEEEtran}
\bibliography{egbib}
}

\end{document}